\documentclass[letterpaper, 10 pt, conference]{ieeeconf}  

\IEEEoverridecommandlockouts                              

\usepackage{graphics} 
\graphicspath{ {./img/} }
\usepackage{svg} 
\usepackage{epsfig} 
\usepackage{subcaption}
\usepackage{adjustbox}
\usepackage{amsmath}

\DeclareMathOperator*{\argmin}{arg\,min}
\usepackage{siunitx}
\usepackage{array}
\usepackage{algorithm2e}
\usepackage{multirow}

\graphicspath{img}

\title{\LARGE \bf
Frequency-based View Selection in Gaussian Splatting Reconstruction
}

\author{Monica M.Q. Li, Pierre-Yves Lajoie, and Giovanni Beltrame
\thanks{*This work was supported by the Natural Sciences and Engineering Research Council of Canada (NSERC) via Vanier Canada Graduate Scholarships. } 
\thanks{The authors are with the Department of Computer and Software Engineering,
        Polytechnique Montréal, Montreal, QC H3T 1J4, Canada
        Corresponding author: {\tt\small monica.li@polymtl.ca}}%
}

\begin{document}
\RestyleAlgo{ruled}

\maketitle
\thispagestyle{empty}
\pagestyle{empty}

\begin{abstract}

Three-dimensional reconstruction is a fundamental problem in robotics perception.
We examine the problem of active view selection to perform 3D Gaussian Splatting reconstructions with as few input images as possible. Although 3D Gaussian Splatting has made significant progress in image rendering and 3D reconstruction, the quality of the reconstruction is strongly impacted by the selection of 2D images and the estimation of camera poses through Structure-from-Motion (SfM) algorithms. Current methods to select views that rely on uncertainties from occlusions, depth ambiguities, or neural network predictions directly are insufficient to handle the issue and struggle to generalize to new scenes. By ranking the potential views in the frequency domain, we are able to effectively estimate the potential information gain of new viewpoints without ground truth data. By overcoming current constraints on model architecture and efficacy, our method achieves state-of-the-art results in view selection, demonstrating its potential for efficient image-based 3D reconstruction.

\end{abstract}

\section{INTRODUCTION}

The next-best-view selection has been a question in active 3D reconstruction for a long time. The aim of selecting the next-best-view is to decide the parts of the scene that need to be discovered and a valid position and orientation to place the camera to view them~\cite{pito_solution_1999}. Careful view selection is important, especially when the cost of gathering or processing additional viewpoints is expensive. For example, when using a drone to map a building, the scanning process is severely time constrained, since it has to be finished before the battery runs out.

As 3D scenes can be presented in various data structures, such as volume element (voxel), surface element (surfel), neural radiance field (NeRF)~\cite{mildenhall2020nerf}, and 3D Gaussians~\cite{kerbl3Dgaussians}, the next-best-view representation needs to be defined based on the data features corresponding to each individual 3D reconstruction method.

\begin{figure*}[htpb]
    \centering
    \includegraphics[width=\textwidth]{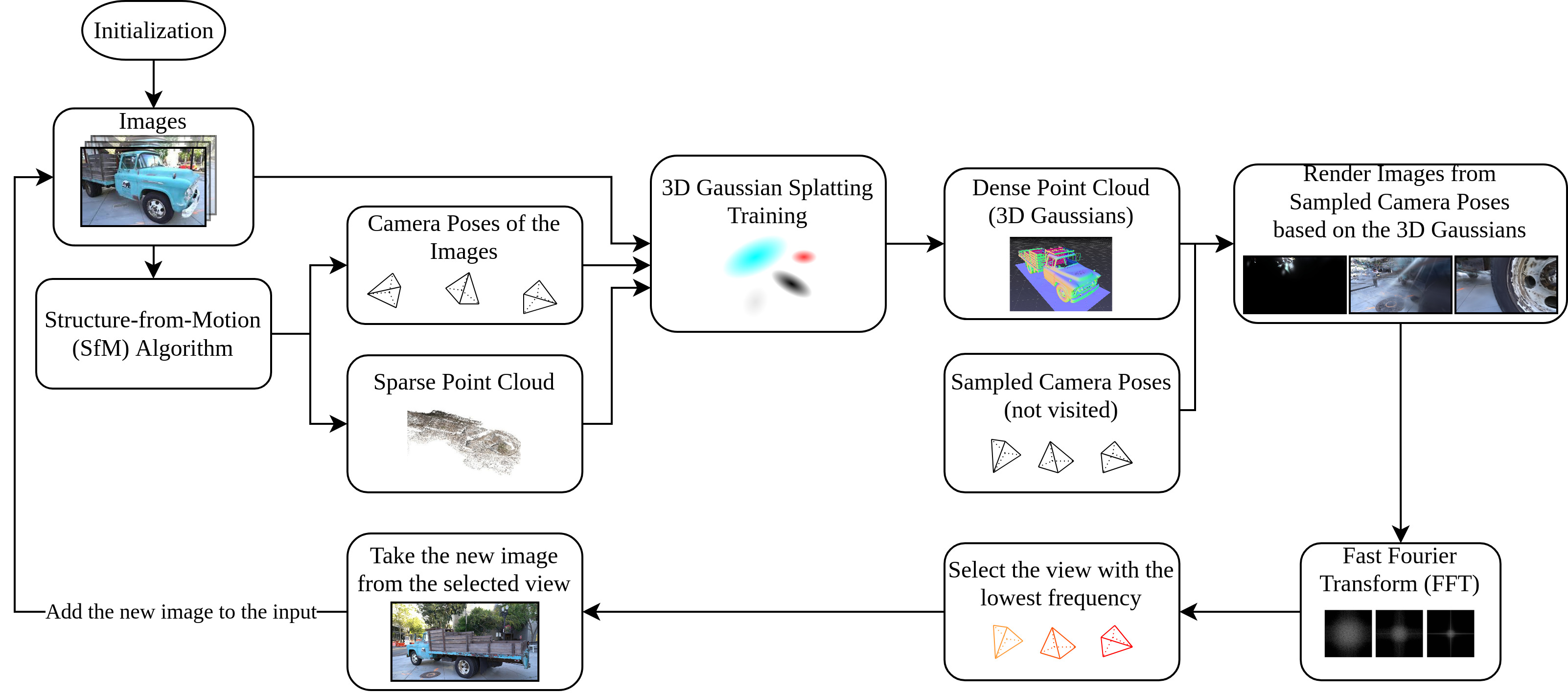}
    \caption{\textbf{The pipeline of the next view selection method}: the scene was initialized with a few images. The images, together with their camera poses and a sparse point cloud generated by SfM, were then used to train a 3D-GS model. The trained 3D-GS model was then used to render images from some sampled camera poses not yet visited. Then, the rendered images were transferred to the frequency domain via FFT. The camera pose with the lowest median frequency would be selected as the next view to visit.}
    \label{pipelne}
\end{figure*}

As shown in Fig.~\ref{pipelne}, our algorithm used the rendering features of Gaussian Splatting models to select views for 3D reconstruction. During the initialization, a few images were taken by the camera as input. The algorithm actively generates the views to visit based on the rendering results from the current reconstructed model. Therefore, the reconstruction can be performed efficiently within a limited number of visited views.

The main contribution of our work is a pipeline for actively selecting the camera view, customized for 3D Gaussian Splatting (3D-GS) models, which utilizes the blur and artifacts of rendered images.
The proposed method achieved reasonable rendering results with only one third of the views in the dataset and significantly reduced the path length between the viewpoints.

\section{RELATED WORKS}

\subsection{Next-best-view Selection}

The concept of next-best-view was first introduced in \cite{connolly_determination_1985}, which designed two algorithms to select the next best range image to visit using partial octree models. This concept was further discussed in active vision and perception \cite{bajcsy_active_1988, aloimonos_active_1988}.

The early application of next-best-view selection is mainly in range scanning and 3D object modeling \cite{pito_solution_1999, wong_next_1999, banta_next-best-view_2000}. Since the scanned objects have both convex and concave localities on their surfaces, the focus of next-best-view was to deal with self-occlusion of the objects \cite{pito_solution_1999, banta_next-best-view_2000} and features of the local contour \cite{chen_vision_2005}.

Later, next-best-view selection was introduced into exploration and navigation. As a result, next-best-view selection methods based on uncertainty and information gain have been developed. For example, \cite{whaite_autonomous_1997} constructed a working autonomous system with a gaze planning strategy based on uncertainty reduction and likelihood maximization. In \cite{wenhardt_active_2007}, the 3D reconstruction was based on probabilistic state estimation with sensor actions, and the next best view was determined by a metric of the state estimation's uncertainty. \cite{gonzalez-banos_navigation_2002} combined the safe region and potential visibility gain to select the region to be explored as a local planner. \cite{dunn_developing_2009} determined the next best view by a cost function, followed by a sensing path for robot motion towards the next best view through a cost-driven recursive search of intermediate viewing configurations.

Recently, the concept of next-best-view has been integrated into various tasks performed by robots. \cite{peraltaNextBestViewPolicy2020} adopted a reinforcement learning approach for scanning buildings with drones, and this work has been extended to a multi-agent setting in \cite{2307.04004}. \cite{breyerClosedLoopNextBestViewPlanning2022} proposed a closed-loop next-best-view planner that drives exploration based on occluded object parts by continuously predicting grasps from the scene reconstruction. \cite{wuPlantPhenotypingDeepLearningBased2019} deployed a similar idea with multiple robot arms for plant phenotyping.

\subsection{Rendering for Novel View Synthesis}

Neural radiance fields, or NeRFs \cite{mildenhall2020nerf}, have been extensively studied to provide innovative vision synthesis using differentiable volume rendering and implicit scene representation. NeRFs can create new views with exceptional multi-view consistency to model 3D scene representations from multi-view 2D photos. Many NeRFs variations have also been developed to handle various challenging scenarios, such as antialiasing \cite{Barron2021MipNeRF3U}, dynamic scenes \cite{du2021nerflow, Gao2021DynamicVS, guo2021adnerf, park2021nerfies, Tretschk2020NonRigidNR}, free camera posture \cite{bian2022nopenerf, Lin_2021_ICCV, meng2021gnerf, zhang2023pose}, and few-shot scenarios \cite{mvsnerf, jain2021putting, niemeyer2022regnerf, yang2023freenerf}. Nevertheless, using NeRFs for new view synthesis frequently results in very long rendering and training times.

To overcome the training time barrier, the 3D-GS \cite{kerbl3Dgaussians} offers a strong substitute for NeRFs by introducing anisotropic 3D Gaussians and effective differentiable splatting, which allows for real-time rendering and quick training while enabling high-quality explicit scene representation. Simultaneous Localization and Mapping (SLAM) research has evolved significantly from hand-crafted methods through the deep learning era to more recent developments focused on rendering view synthesis of 3D scene radiance, thanks to recent advancements in these new rendering methods \cite{tosi2024nerfs}. Nevertheless, blur and artifacts are frequently introduced into the generated images due to the over-reconstruction of 3D Gaussians during Gaussian densification. Our method utilized these blur and artifacts to select the views to be visited based on their distinguishing features in the frequency space.

\subsection{Active view planning in Novel View Synthesis Context}

In the setting of Neural Radiance Fields (NeRF), some view planning methods based on uncertainties have been proposed. Jin et al.\cite{jin_neu-nbv_2023} trained a neural network for uncertainty rendering to select the next-best-view. Their approach incorporates position encoding, multi-layer perceptrons for feature extraction--similar to PixelNeRF\cite{yu2021pixelnerf}--and an LSTM module~\cite{10.1162/neco.1997.9.8.1735} to predict the jumping distance in the neural radiance field. However, this supervised learning model struggles to generalize across scenes with varying scales, such as when reconstructing a household object compared to an architectural structure. Similarly, \cite{xueNeuralVisibilityField2024} trained a model to predict the uncertainty for NeRF rendering with a focus on occlusion. Since the uncertainty is calculated by accumulating the rays in the radiance field, it cannot be applied to Gaussian splatting models as there are no tracing rays.

Jiang et al.~\cite{Jiang2023FisherRF} quantified observed information gain within Radiance Fields by leveraging Fisher Information, and the authors applied this concept for navigation in their subsequent work~\cite{liu2024beyond}. Although their information gain was implemented onto Gaussian splatting, the camera poses were known in advance without SfM. The poses are generally not known during the image acquisition stage and the Fisher information gain needs to be traced via the rays in the neural radiance fields, making the method unsuitable for 3D reconstruction relying solely on image input.

Jin et al.\cite{jin2024gs} adopted 3D-GS in their active planning framework for view planning. In their work, the 3D Gaussians mainly act as a provider for geometry and unobserved regions, which were input for a traditional view planner to establish goals for exploration, by integrating the information gain of each pixel in the potential view frustum. As this method only used the point cloud generated by Gaussian splatting as input for the planner and was not optimized for the features of 3D Gaussian models, it served better as a planner for navigation rather than for 3D reconstruction.

\section{METHOD}

\subsection{Preliminaries}

\subsubsection{Structure-from-Motion(SfM)}

The input to our algorithm is a set of images collected from a static scene, visited by the camera at the initialization stage. A Structure-from-Motion(SfM) algorithm is applied to compute the camera pose for each image, along with a generated sparse point cloud~\cite{7780814}. The original Gaussian Splatting used COLMAP for SfM~\cite{schoenberger2016sfm}, which involves three phases: feature extraction, feature matching, and mapping. As COLMAP's mapping is not accurate when the input contains only a few images at initialization, we used GLOMAP~\cite{pan2024glomap}, a more efficient and scalable mapper published by the same research team for the reconstruction process.

\subsubsection{3D Gaussian Splatting}

Based on the input images, camera poses, and sparse point cloud, the 3D Gaussian Splatting Algorithm (3D-GS)~\cite{kerbl3Dgaussians} created a set of 3D Gaussians, defined by a position (the mean of $X, Y, Z$ coordinates of each Gaussian), covariance matrix, and opacity. This results in a compact representation of the 3D scene.

Once the model has been trained, camera poses near the current camera pose were sampled. For each sampled camera pose, we splatted the volumetric Gaussians in the trained model sequentially onto the camera plane to obtain the rendered image. In our current setting, the camera pose sampled from the dataset needs to be transformed into the current world space. The method is introduced below.

\subsubsection{Kabsch-Umeyama algorithm}

As the input images were photos taken by a monocular camera and there is no depth information, the camera poses of the input images generated by the SfM algorithm need a rigid transformation to match the camera poses of the same images in the dataset. 
We chose the Kabsch-Umeyama registration algorithm~\cite{umeyama1991}, which is widely used for point-set registration in computer graphics.
This algorithm determines the ideal rotation matrix by minimizing the root mean squared deviation (RMSD) between two matched sets of points~\cite{umeyama1991}. 
The algorithm provides the rotation matrix, scaling factor, and translation vector to transform the camera poses from the dataset to the world space where the model is to be trained.


\subsubsection{Fourier Transform of images}

The Fourier Transform is an important image processing tool used to decompose an image into its sine and cosine components~\cite{rauh2013sparse}. The output of the transformation represents the image in the Fourier or frequency domain, while the input image is the spatial domain equivalent. In the Fourier domain image, each point represents a particular frequency contained in the spatial domain image.

\begin{figure}[htpb]
    \centering
    \begin{subfigure}[htpb]{\columnwidth}
        \centering
        \includegraphics[width=\columnwidth]{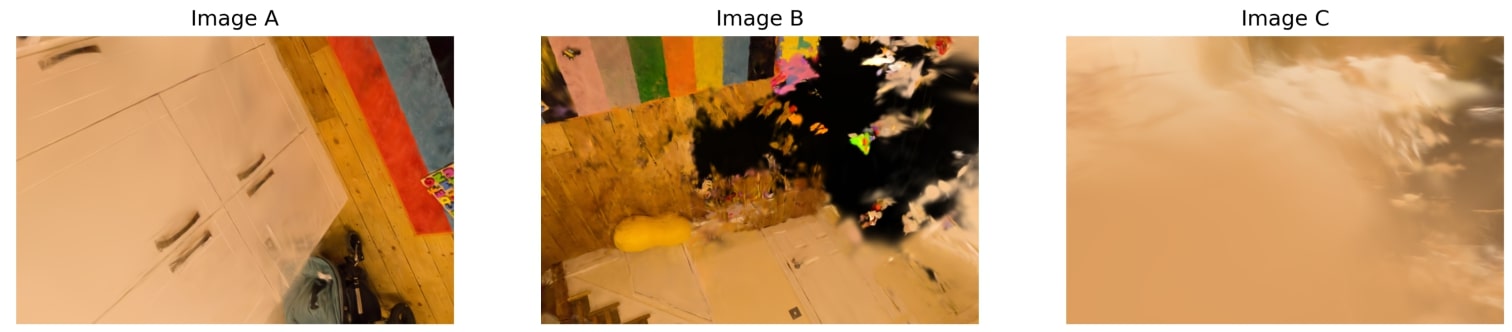}
        \caption{Rendered images from three views}
        \label{fft_input}
    \end{subfigure}
    \begin{subfigure}[htpb]{1.08\columnwidth}
        \centering
        \hspace{-0.08\columnwidth}
        \includegraphics[width=\columnwidth]{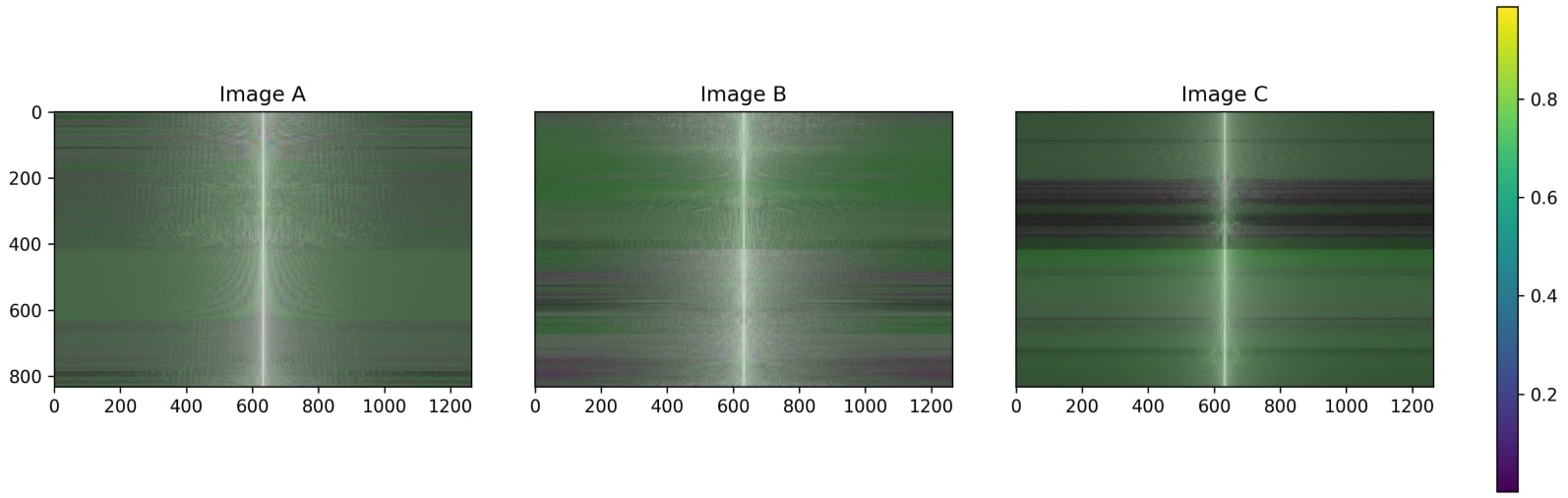}
        \caption{The normalized magnitude spectrum of the image frequencies}
        \label{fft_specturm}
    \end{subfigure}
    \begin{subfigure}[htpb]{\columnwidth}
        \centering
        \includegraphics[width=\columnwidth]{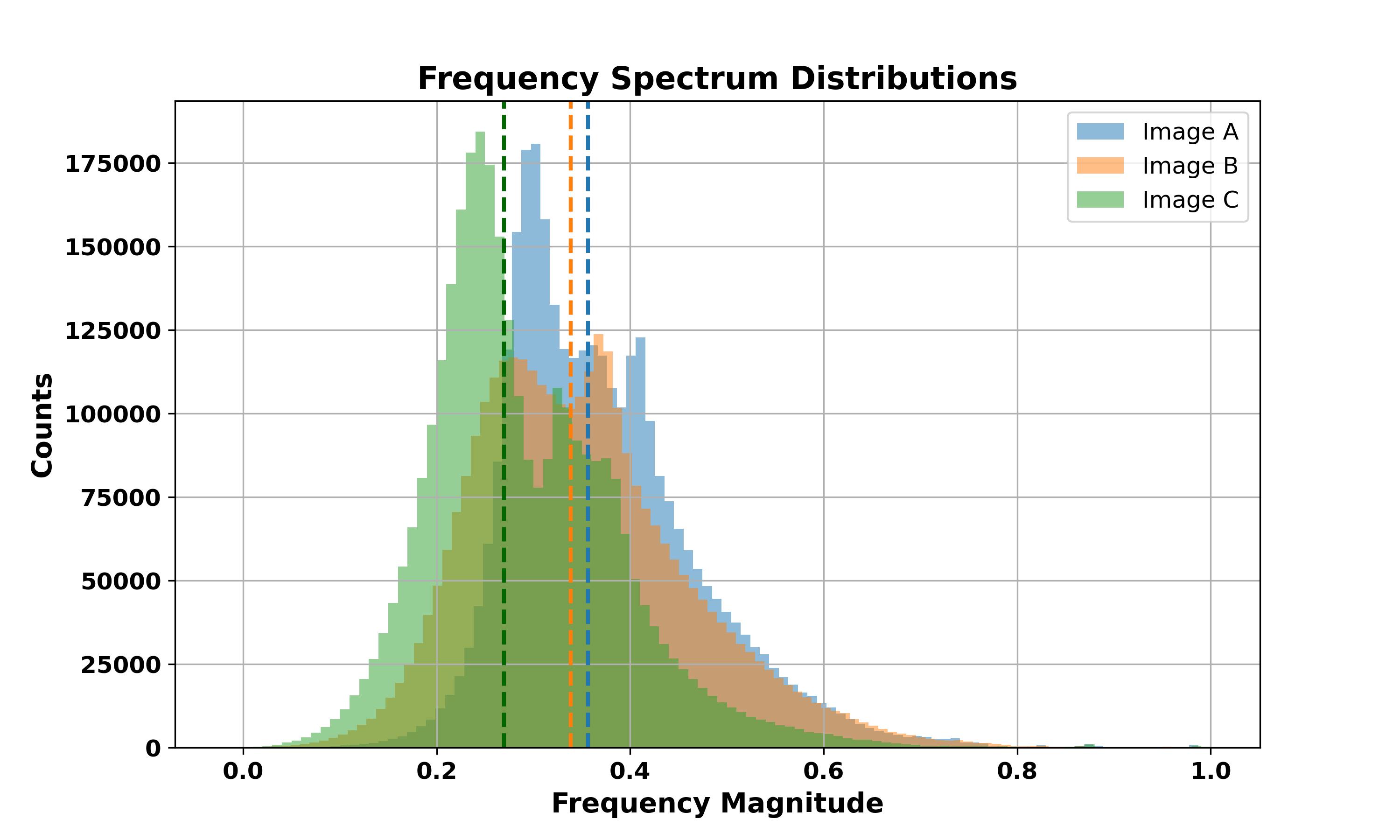}
        \caption{The distribution of frequencies in the magnitude spectrum, with the median of each distribution marked with dashed line}
        \label{fft_bin}
    \end{subfigure}
    \caption{\textbf{FFT of rendered images}: The blur and artifacts of poorly rendered images were converted into low frequency signals and therefore the view with low frequency signals could be selected as the view to visit next.}
    \label{fft_presentation}
\end{figure}

As illustrated in Fig.~\ref{fft_presentation}, the rendered images in Fig.~\ref{fft_input} are transformed into the frequency domain in Fig.~\ref{fft_specturm}, where the zero values were shifted to the center of the image for observation purposes. We observe that the artifacts and blur in the rendered image are converted to low-frequency signals in the spectrum. Therefore, a poorly rendered image would have more low-frequency signals in its spectrum. The frequency distributions and the median of each spectrum are shown in Fig.~\ref{fft_bin}. As the median of Image C's frequency is the lowest, indicating more rendering artifacts and blur, the camera view of Image C would be selected as the next view.

\subsection{Algorithm}

Our proposed algorithm is represented in Alg.~\ref{alg:nbv}. The scene is initialized with a few input images. The camera poses of the images are generated via GLOMAP~\cite{pan2024glomap}. We then trained a 3D Gaussian model based on the images, camera poses, and the sparse point cloud. The 3D Gaussian model is then used for rendering from the candidate views.

The candidate views are sampled from the yet-to-be-visited views in the dataset if they were within the distance threshold $r$ to the current view in Euclidean space. In the real world, the distance threshold should be selected based on the distance the robot can move in the next time step, as well as the surrounding obstacles in the environment. Before the rendering process, they are transformed to the current world frame via the Kabsch-Umeyama algorithm.

The rendered images are then transformed to the frequency domain via Fast Fourier Transform (FFT), and the view with the rendered image having the lowest median frequency is selected as the next view to visit. The view is then marked as visited, and its corresponding image added to the input images for training in the next iteration.

\begin{algorithm}
\caption{Next-best-view selection with a dataset}\label{alg:nbv}
\KwData{$A = \{ a_i | i \in [1, \cdots n] \}$: Dataset of images, the maximum of iterations number $l$, the distance threshold $r$ to sample close views}
\KwResult{Selected views $B$}
 Initialization: use SfM to generate camera pose set $P_0 = \{p_{0_i}|  i \in [1, \cdots n] \}$ and a sparse point cloud $S_0$. Choose the first $m$ images in $A$ as the first visited images set $V$. Set the current pose $p_{c_0} = p_m$. Set function $FFT$ be the 2D fast Fourier transform. The set of views to be visited $B = \emptyset$\; 
 \While{$m \leq l$}{
  Use SfM to generate camera pose set $P_c = \{p_i|  i \in [1, \cdots m] \}$ and a sparse point cloud $S_c$\;
  $P_d = \emptyset$\;

  \For{$v_i \in V$}{
    $P_d.$insert$(p_{0_i} \in P_0$)\;
    }

  Use Kabsch-Umeyama algorithm to get the rigid transformation (Rotation $R$, Scaling $s$, Translation $t$) from the positions in $P_d$ to the positions in $P_c$ \;
  Use $V$, $P_c$ and $S_c$ to train the 3D Gaussian model $G_c$ to represent the observed scene\;
  $F = \emptyset$ \;
  \For{$p_j \in P_0$}
    {
      {\eIf{$\|p_j - p_{c_0} \| \leq r$ and $a_j \notin V$}
        {$p_{c_j} = R \times s \cdot p_j + t$\;
         Render Image $i_j$ from $p_{c_j}$ on $G_c$\;
         $f_j$ = median$(FFT(i_j))$\;
         $F.$insert$(f_j)$\;
        }
        {continue\;}
      $j = \argmin(f_j \in F)$\;
      $B.$insert$(p_{0_j} \in P)$\;
      $V.$insert$(a_{j} \in A)$\;
      $m++\;$
      }
    }
}
\end{algorithm}

\section{EXPERIMENTS}


\subsection{Implementation Details}

Our method can be applied to 3D-GS rendering models. We implemented the computation of Fast Fourier Transform on rendered images with CUDA-accelerated PyTorch. The experiments were performed on a workstation with a 13th Gen Intel(R) Core(TM) i5-13600K CPU and an NVIDIA GeForce RTX 4070 graphic card. The spherical degree for the Gaussians was set to $0$ to accelerate the training speed. The initialization iteration was set to $1000$ and incremented by $100$ when a new view was selected. The training ended when $90$ more views were added, resulting in $10{,}000$ iterations at the end of the training.

In the read world, the selection of next views occurs during the image acquisition process while the reconstruction occurs after all the images have been acquired. Therefore we applied a coarse and a fine setting for the training of the 3D Gaussians. The coarse training set the spherical harmonics of the 3D Gaussians to zero with a maximum training iteration of $10,000$, as the detailed textures are less important during the view selection. The fine training set the spherical harmonics of the 3D Gaussians to three with a maximum training iteration of $30,000$, as the default setting of 3D-GS.



\subsection{Datasets}

For training and testing, we use the same datasets used in the original 3D-GS paper~\cite{kerbl20233d} and conduct experiments on images from a total of four real scenes. Specifically, our approach is extensively evaluated on two common benchmark datasets: Playroom and Dr. Johnson from the Deep Blending dataset~\cite{10.1145/3272127.3275084} and Truck and Train from Tanks \& Temples \cite{10.1145/3072959.3073599}. The selected scenes exhibit diverse styles, ranging from bounded indoor environments to unbounded outdoor ones. For each scene, the first ten images are taken as input to simulate the initial movement of a camera in the real world. To divide the datasets into training and test sets, we follow the approach of 3D-GS and allocate every 8th photo to the test set. The resolution of the final rendered images is the same as in 3D-GS as well.

\subsection{Metrics and Baselines}

Similar to the 3D-GS, our evaluations use image quality metrics such as peak signal-to-noise ratio (PSNR), structural similarity index (SSIM) and learned perceptual image patch similarity(LPIPS). Additionally, we cover the number of views visited as well as the distances to visit all the views.

We quantitatively and qualitatively compare our method against the original 3D-GS dataset as a baseline. 

\section{RESULTS}

\begin{figure}[htpb]
    \vspace{-2em}
    \centering
    \begin{subfigure}[ht]{\columnwidth}
        \centering
        \includegraphics[width=0.7\columnwidth]{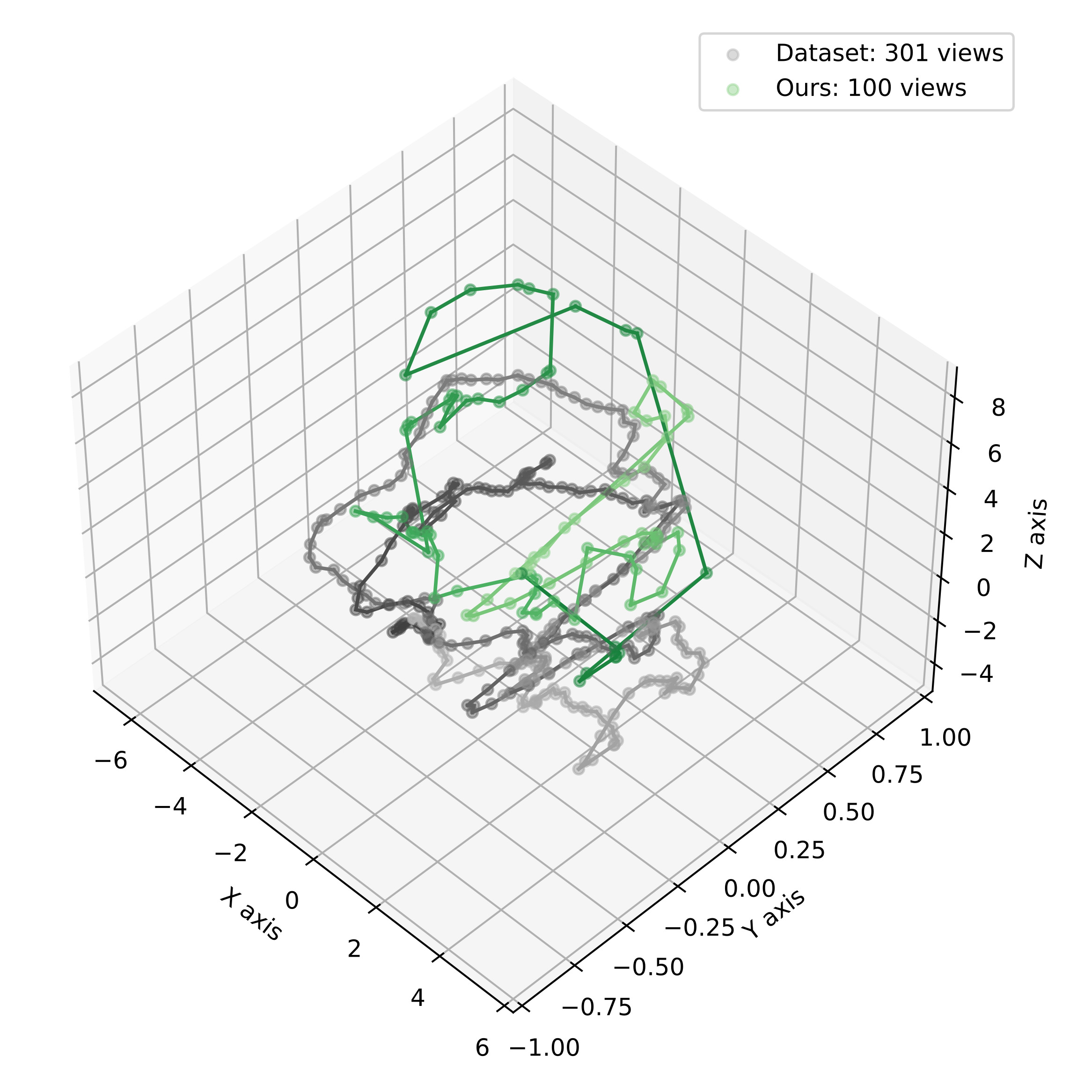}
    \end{subfigure}%
    \vspace{-0.5em} 
    \begin{subfigure}[ht]{\columnwidth}
        \centering
        \includegraphics[width=0.7\columnwidth]{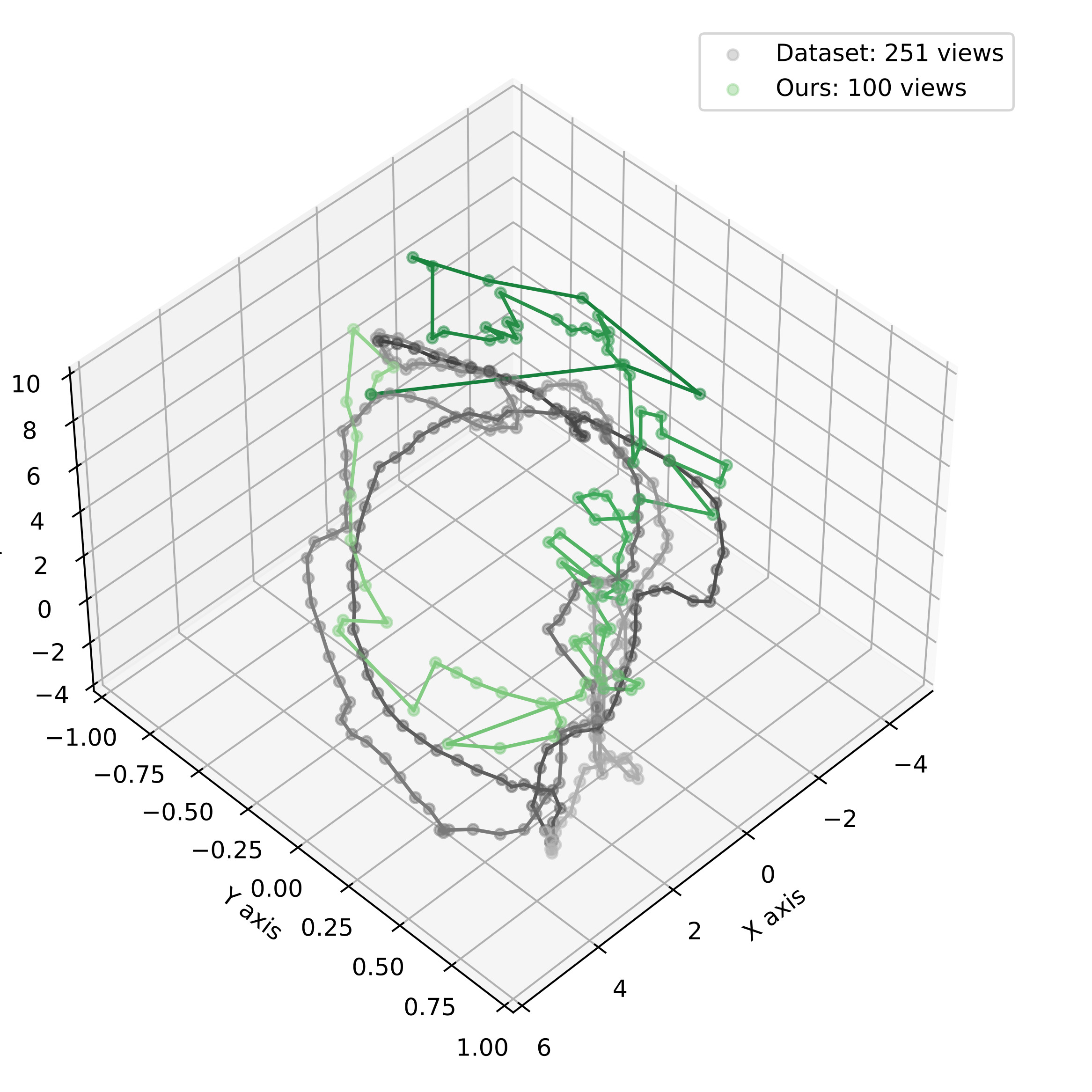}
    \end{subfigure}%
    \vspace{-0.5em} 
    \begin{subfigure}[ht]{\columnwidth}
        \centering
        \includegraphics[width=0.7\columnwidth]{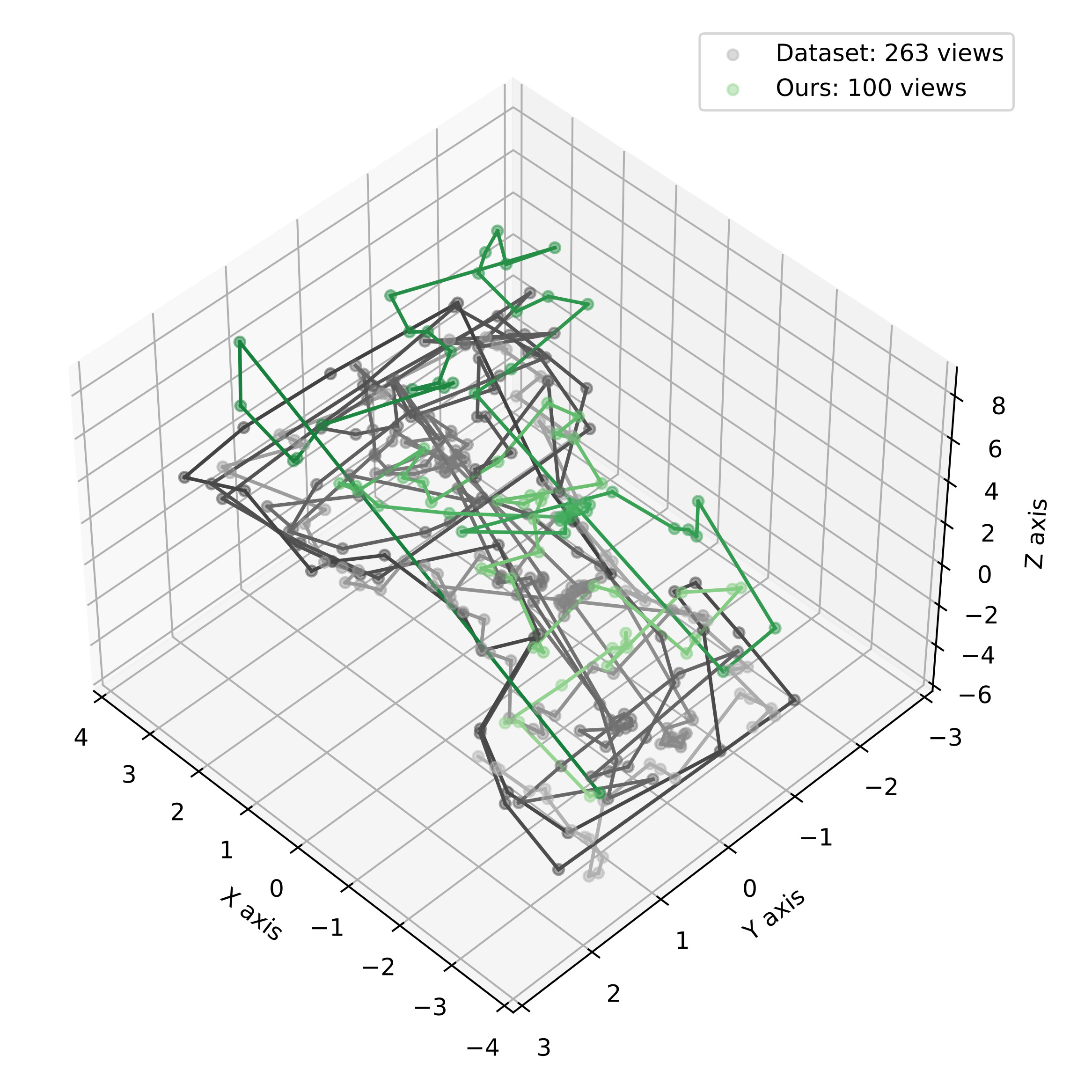}
    \end{subfigure}%
    \vspace{-0.5em} 
    \begin{subfigure}[ht]{\columnwidth}
        \centering
        \includegraphics[width=0.7\columnwidth]{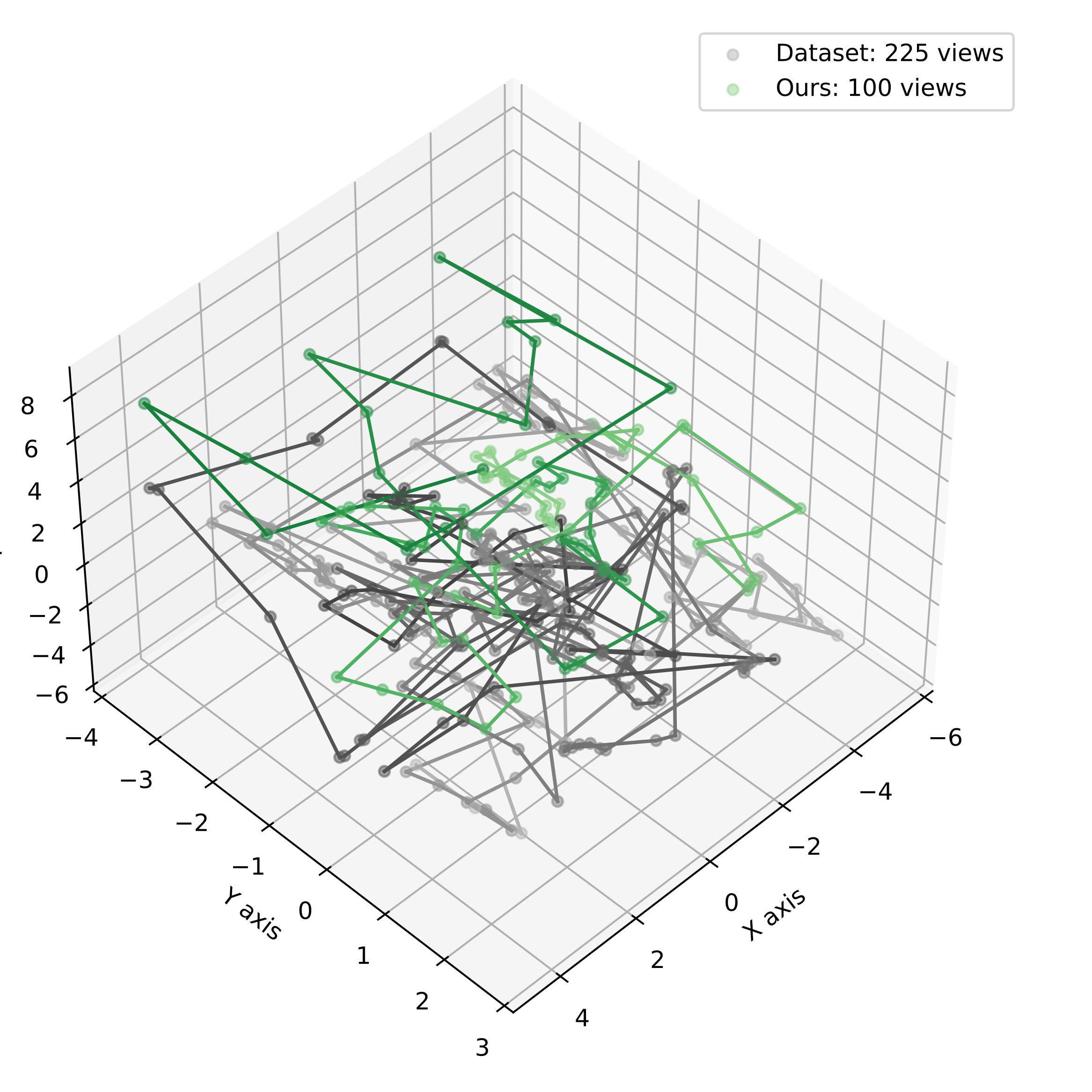}
    \end{subfigure}
    \vspace{-0.5em}
    \caption{\textbf{Trajectories to visit all views and our selected views}: our selected views are adjusted 4 units above along z-axis for presentation purpose. From top to bottom are individual scenes: Train, Truck, Dr Johnson and Playroom.}
    \label{trajectories}
\end{figure}

The trajectories of the views are shown in Fig.~\ref{trajectories} and the distances to visit the views are shown in Table~\ref{dist}. We selected 100 views while the amount of all views in dataset arranges from 250 to 300. The trajectories to visit all the views are reduced to between 30\% and 25\% of the original trajectories' length.

As shown in Fig.~\ref{trajectories}, the significant reduction in the length of trajectories is due to the fact that the active view selection can finish the scanning in just one revolution around the object to be reconstructed, where as two to three revolutions are needed using the approach in the dataset.

\begin{table}[h]
\caption{Traveling distances to visit the selected views}
\label{dist}
\vspace{-0.5em}
\begin{center}
\begin{tabular}{|c|S[table-format=3.2]|S[table-format=3.2]|S[table-format=3.2]|S[table-format=3.2]|}
\hline
 & {Train} & {Truck} & {Dr Johnson} & {Playroom} \\
\hline
Dataset & 332.39 & 406.72 & 387.35 & 455.68 \\
\hline
Ours & 60.16 & 104.74 & 77.05 & 104.75 \\
\hline
\end{tabular}
\end{center}
\vspace{-0.5em}
\end{table}

\begin{figure*}[htbp]
    \centering
    \makebox[\textwidth][c]{
    \begin{tabular}{@{}c@{}c@{}c@{}c@{}} 
        \adjustbox{valign=m}{\includegraphics[width=0.24\textwidth]{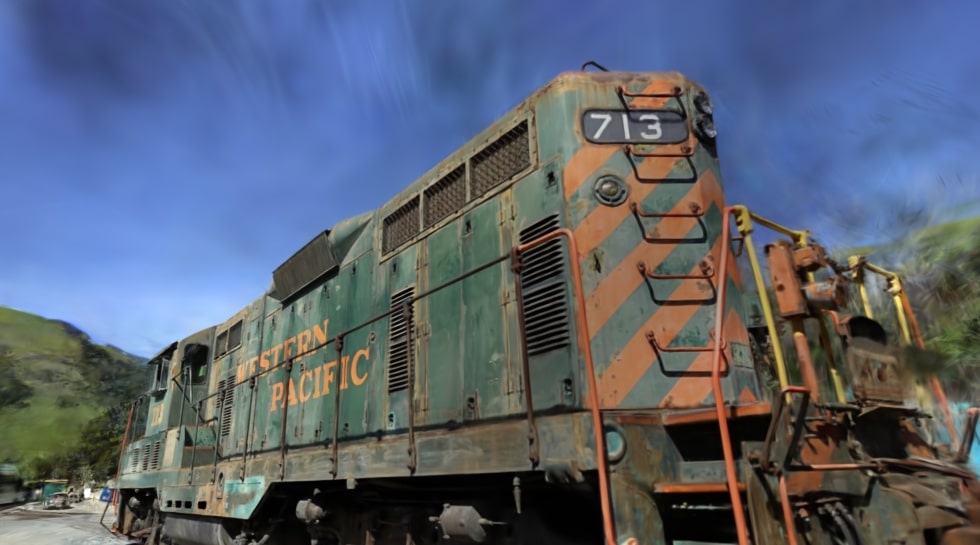}} &
        \adjustbox{valign=m}{\includegraphics[width=0.24\textwidth]{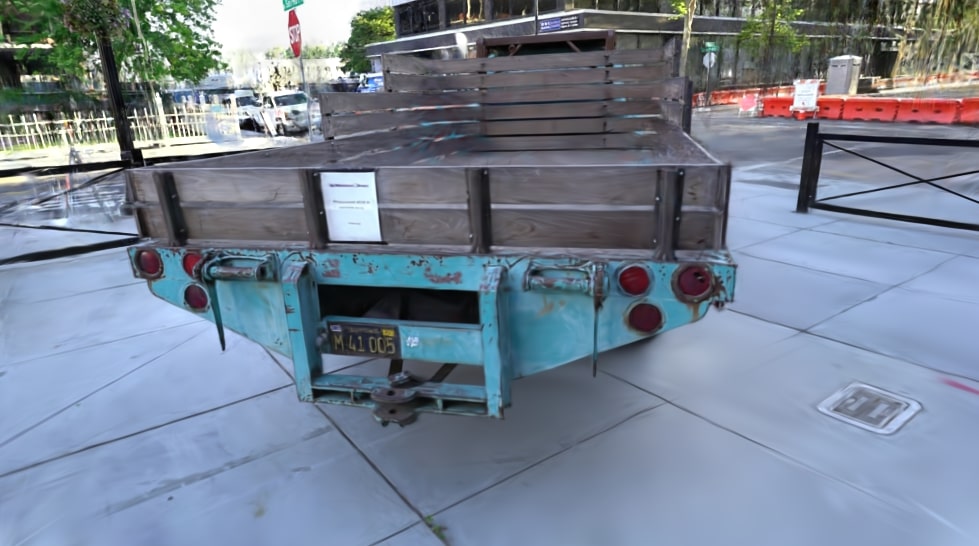}} &
        \adjustbox{valign=m}{\includegraphics[width=0.24\textwidth]{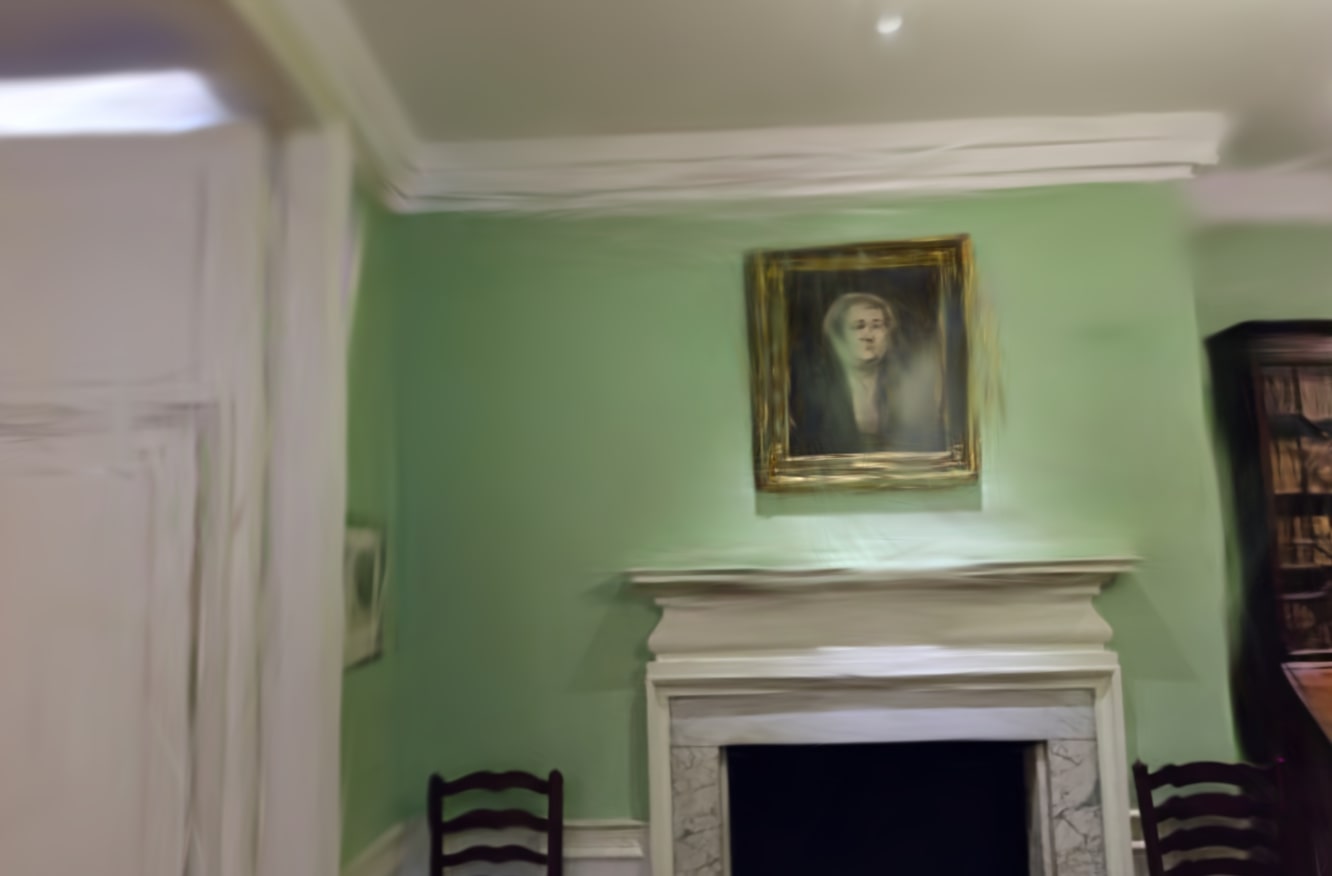}} &
        \adjustbox{valign=m}{\includegraphics[width=0.24\textwidth]{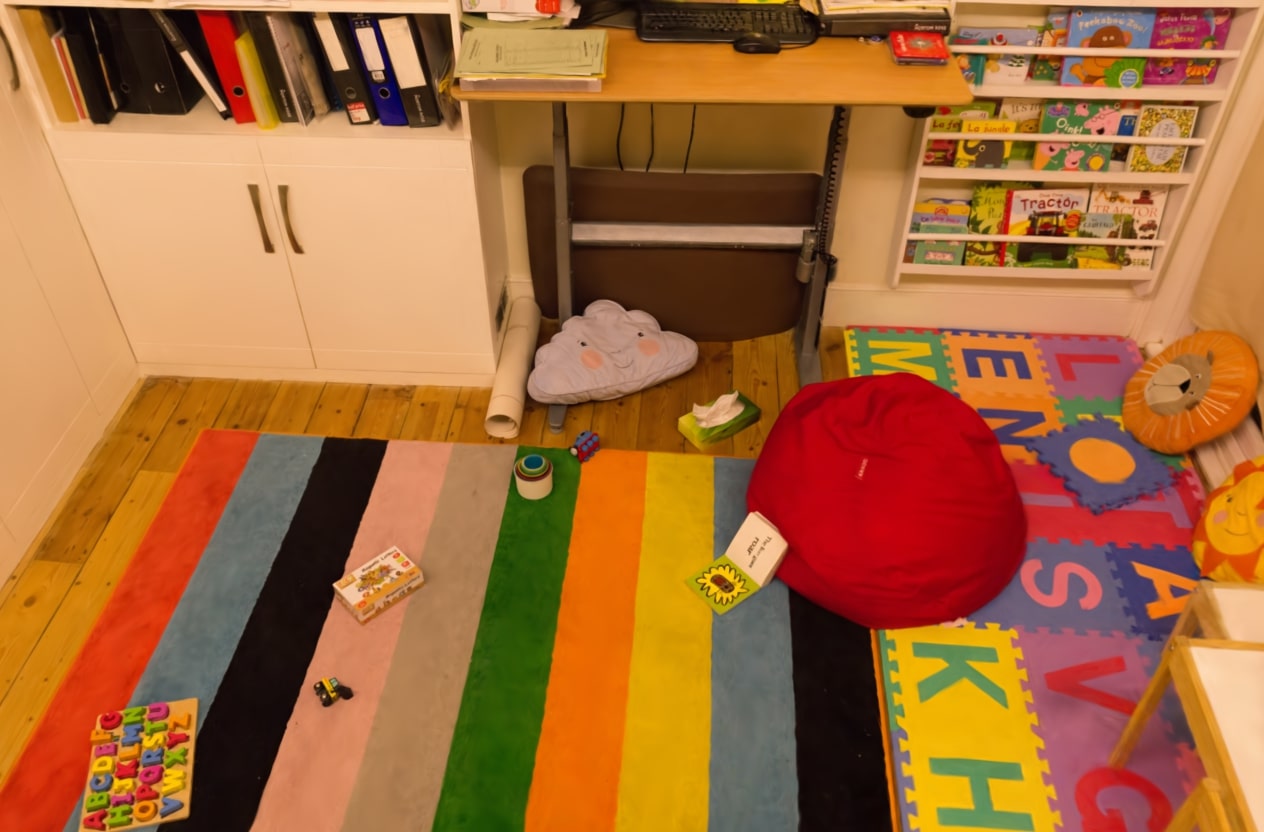}} \\
        \adjustbox{valign=m}{\includegraphics[width=0.24\textwidth]{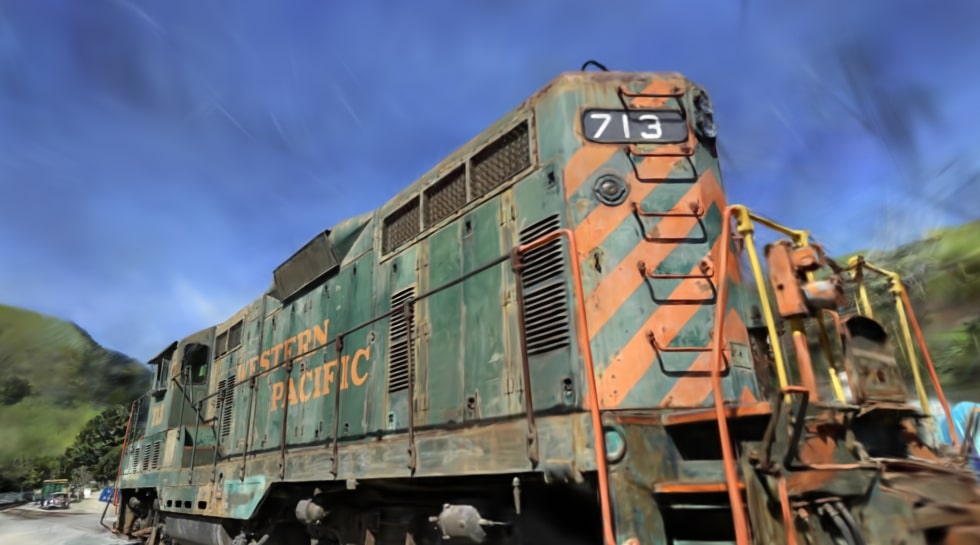}} &
        \adjustbox{valign=m}{\includegraphics[width=0.24\textwidth]{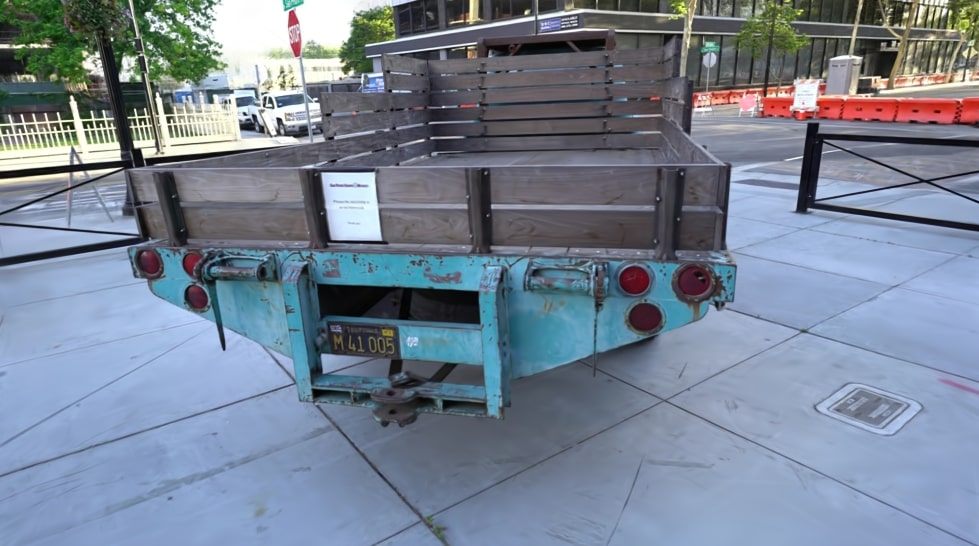}} &
        \adjustbox{valign=m}{\includegraphics[width=0.24\textwidth]{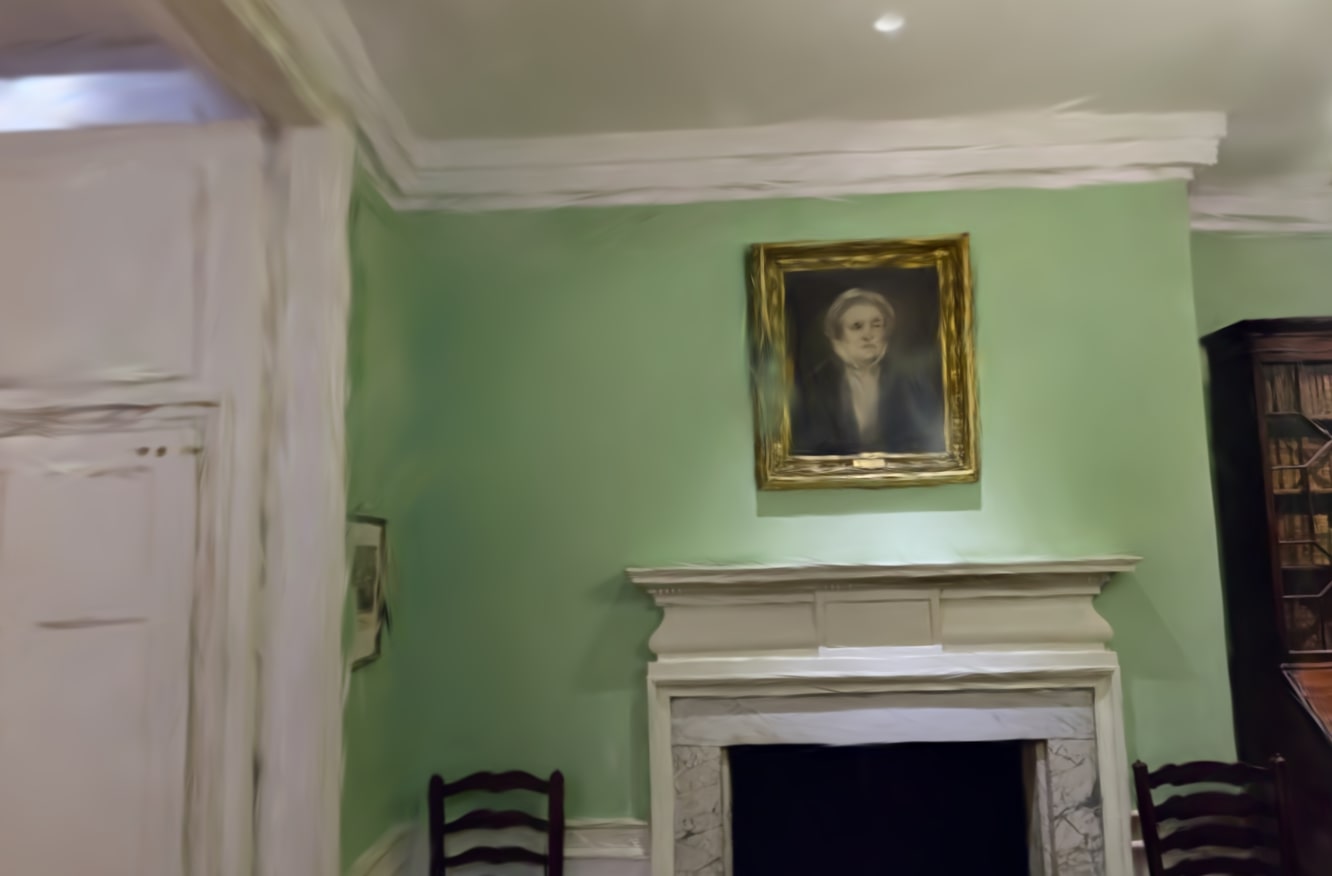}} &
        \adjustbox{valign=m}{\includegraphics[width=0.24\textwidth]{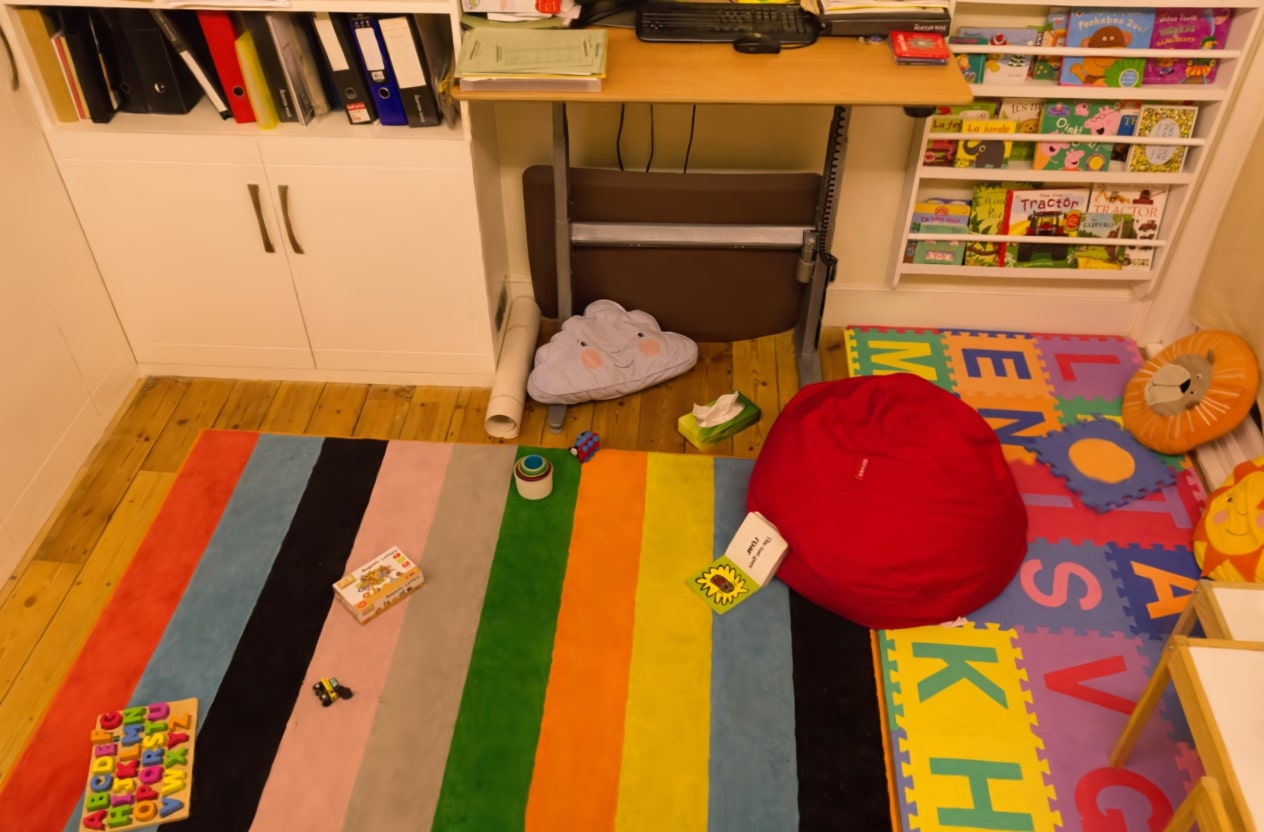}} \\
        \adjustbox{valign=m}{\includegraphics[width=0.24\textwidth]{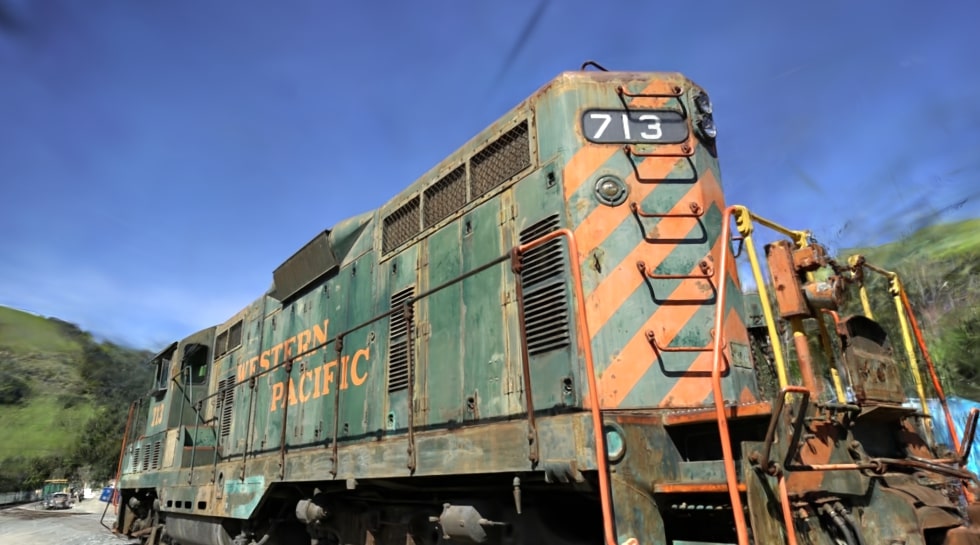}} &
        \adjustbox{valign=m}{\includegraphics[width=0.24\textwidth]{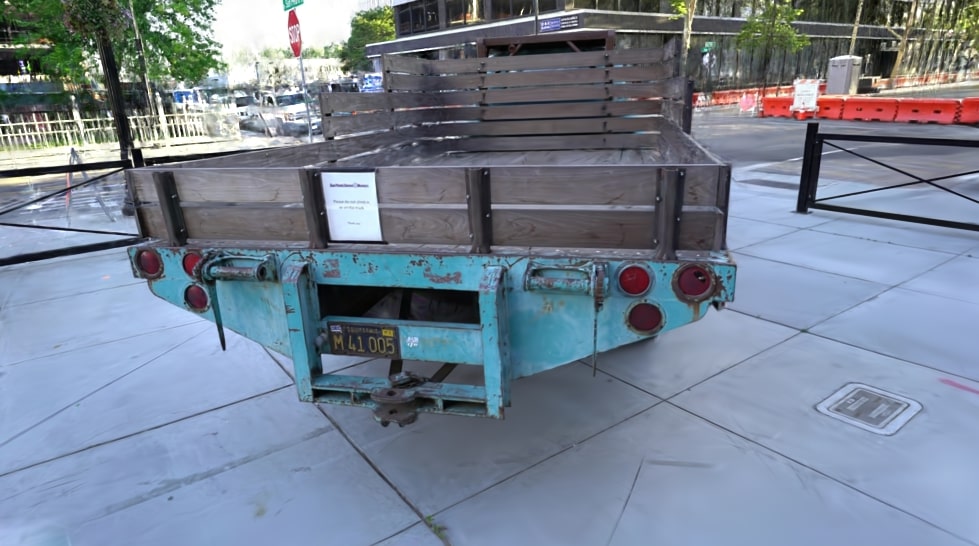}} &
        \adjustbox{valign=m}{\includegraphics[width=0.24\textwidth]{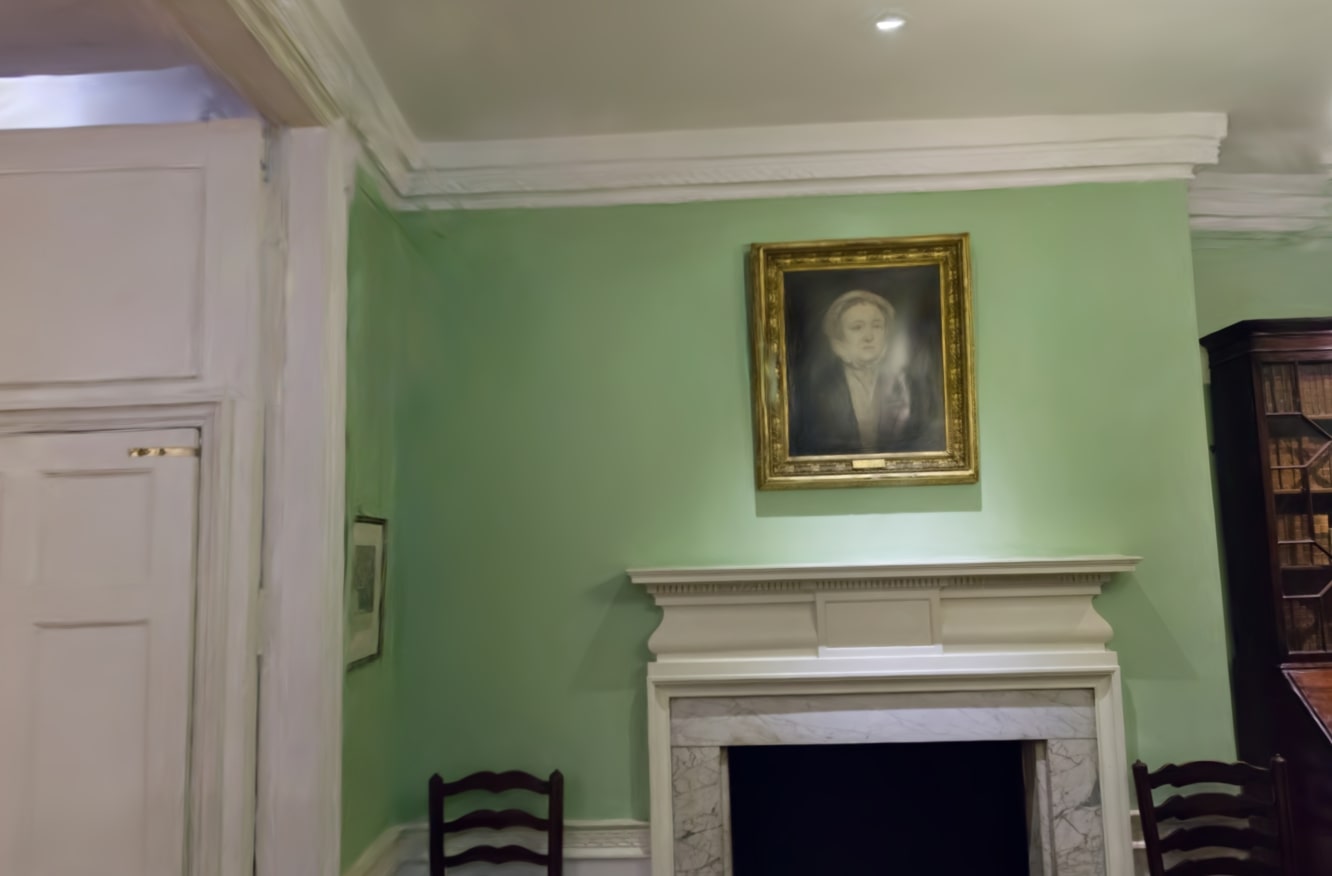}} &
        \adjustbox{valign=m}{\includegraphics[width=0.24\textwidth]{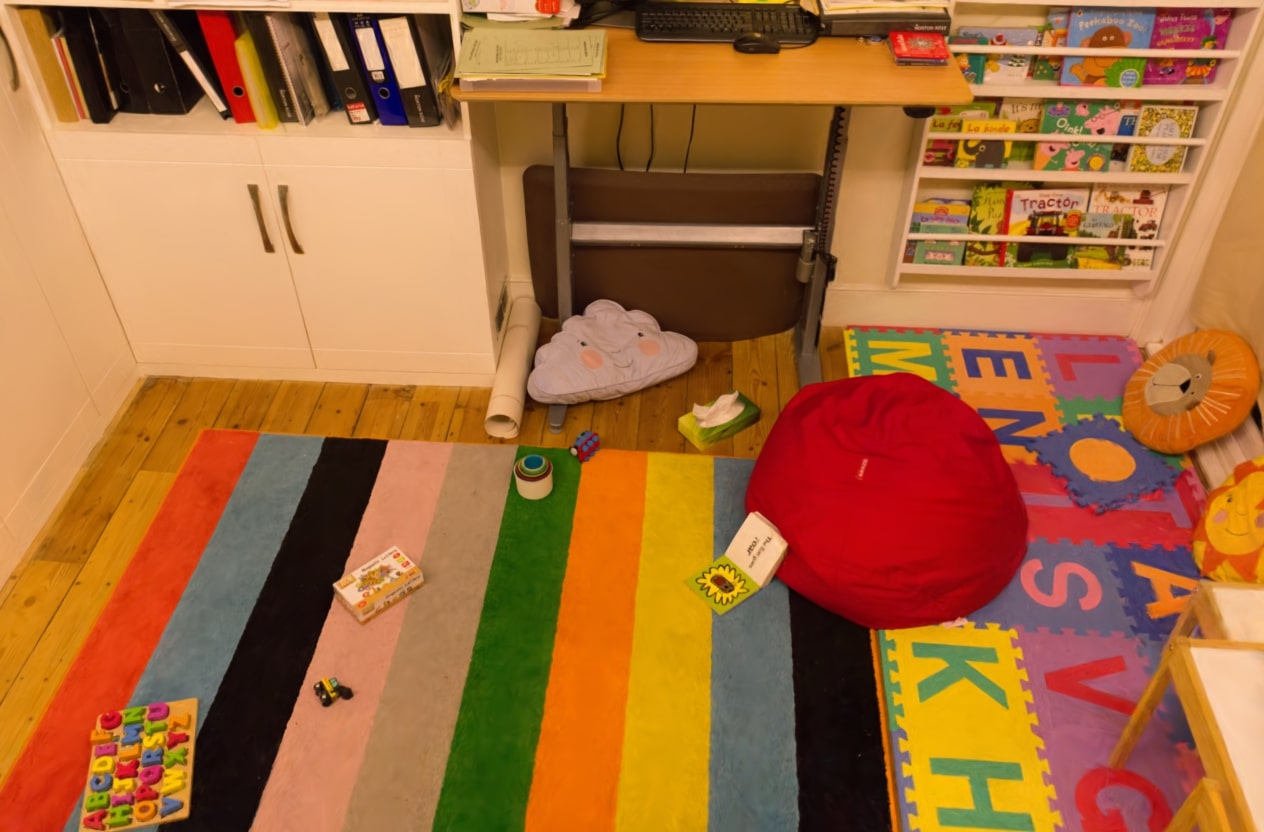}} \\
        \adjustbox{valign=m}{\includegraphics[width=0.24\textwidth]{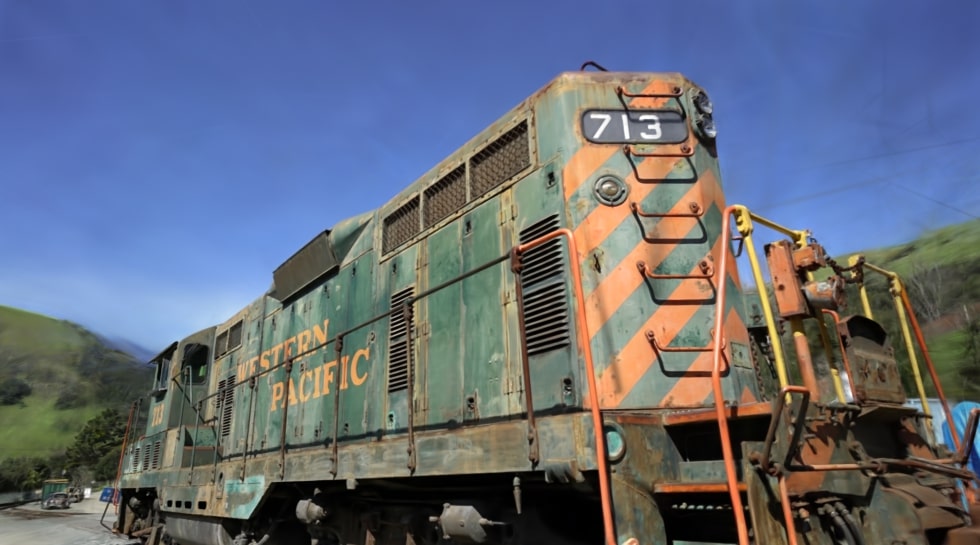}} &
        \adjustbox{valign=m}{\includegraphics[width=0.24\textwidth]{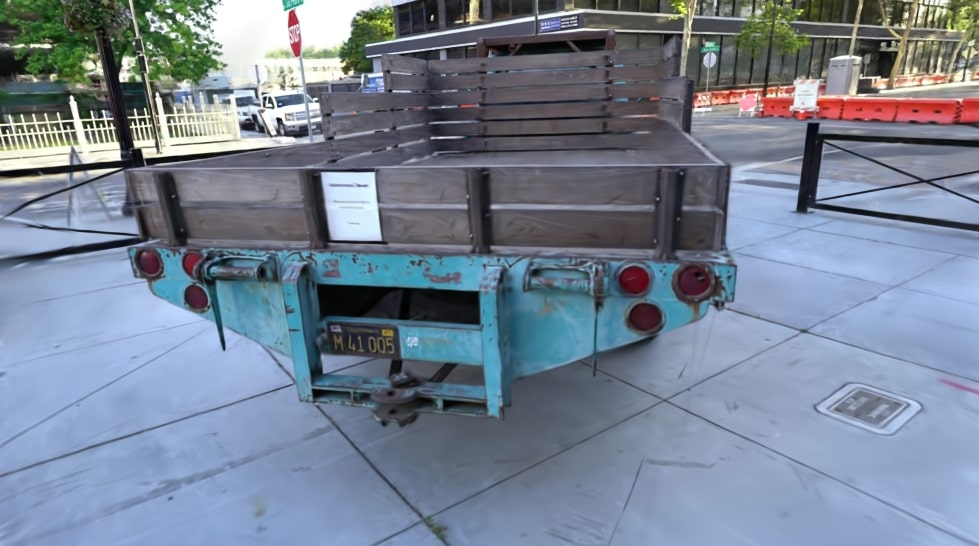}} &
        \adjustbox{valign=m}{\includegraphics[width=0.24\textwidth]{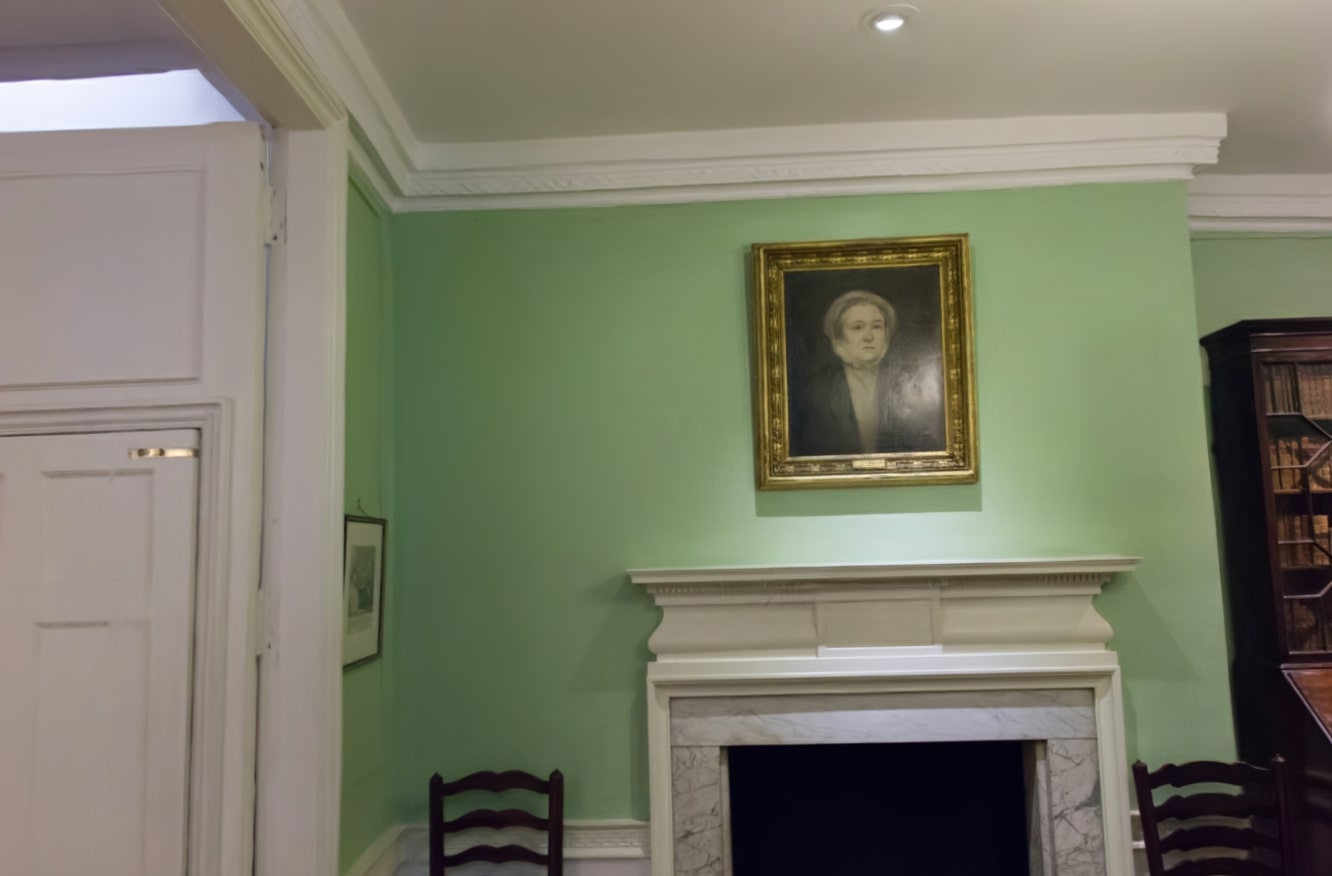}} &
        \adjustbox{valign=m}{\includegraphics[width=0.24\textwidth]{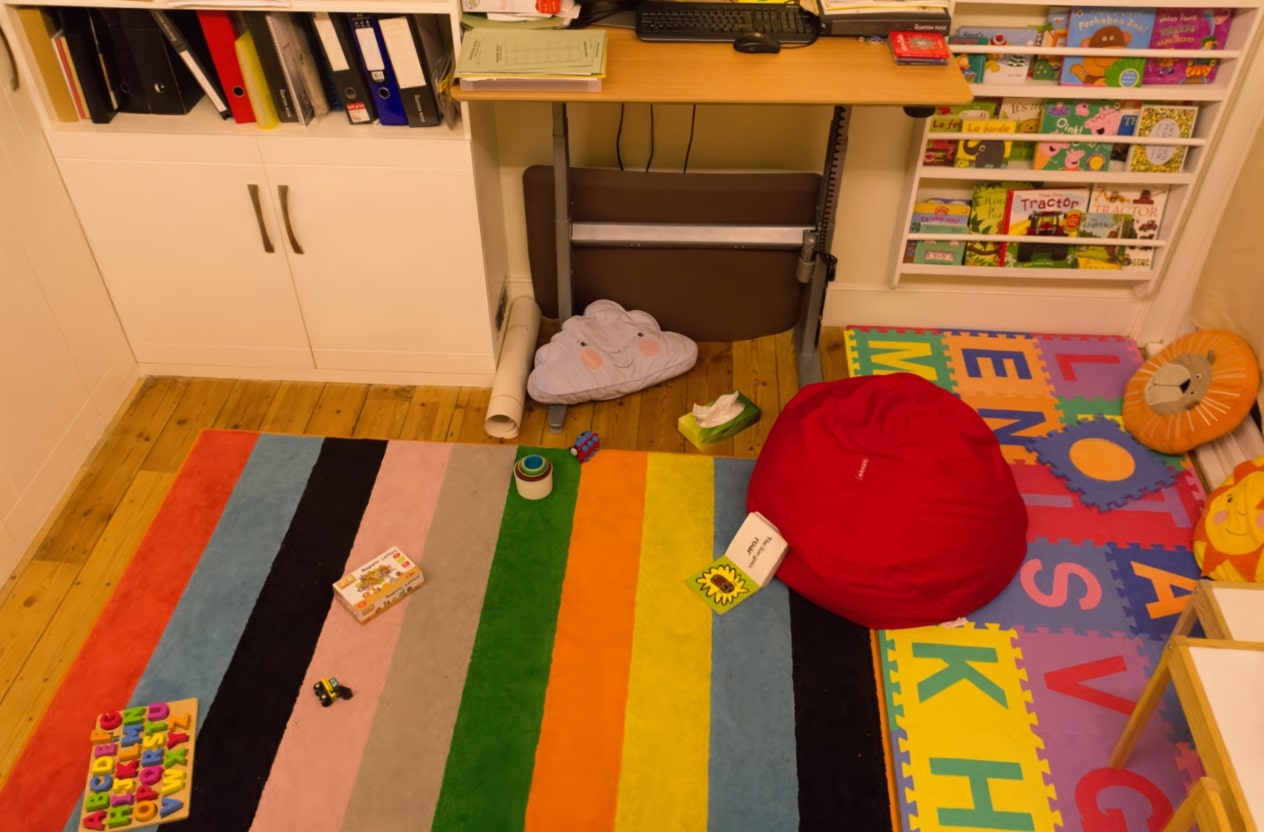}} \\
        \adjustbox{valign=m}{\includegraphics[width=0.24\textwidth]{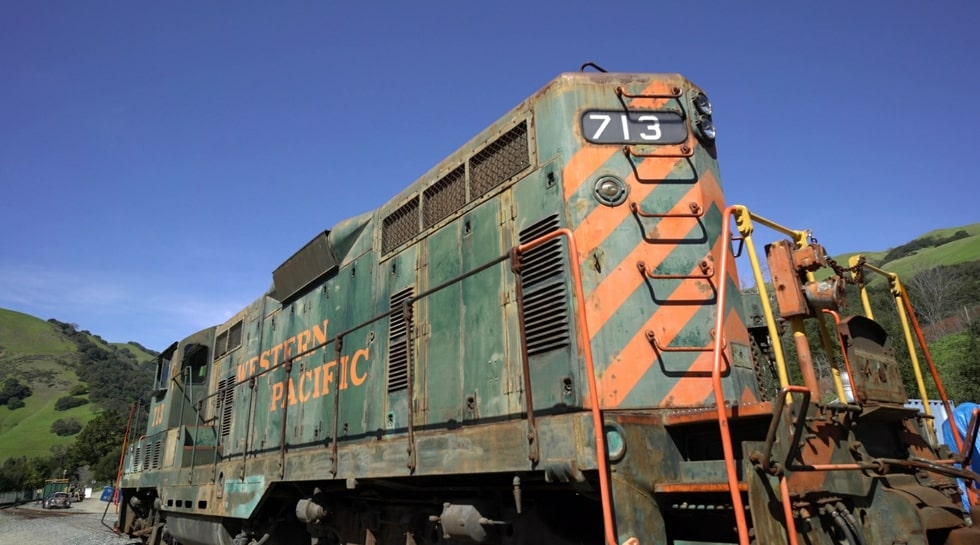}} &
        \adjustbox{valign=m}{\includegraphics[width=0.24\textwidth]{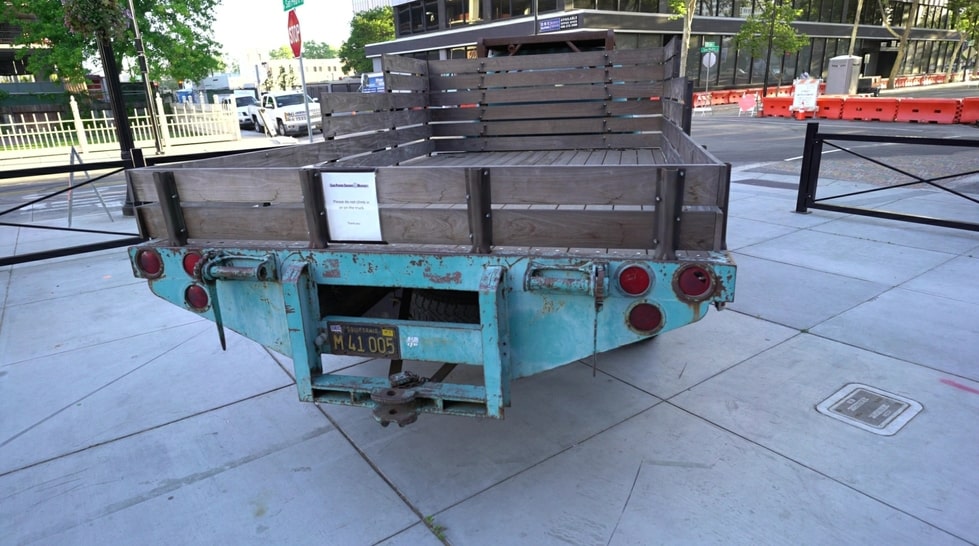}} &
        \adjustbox{valign=m}{\includegraphics[width=0.24\textwidth]{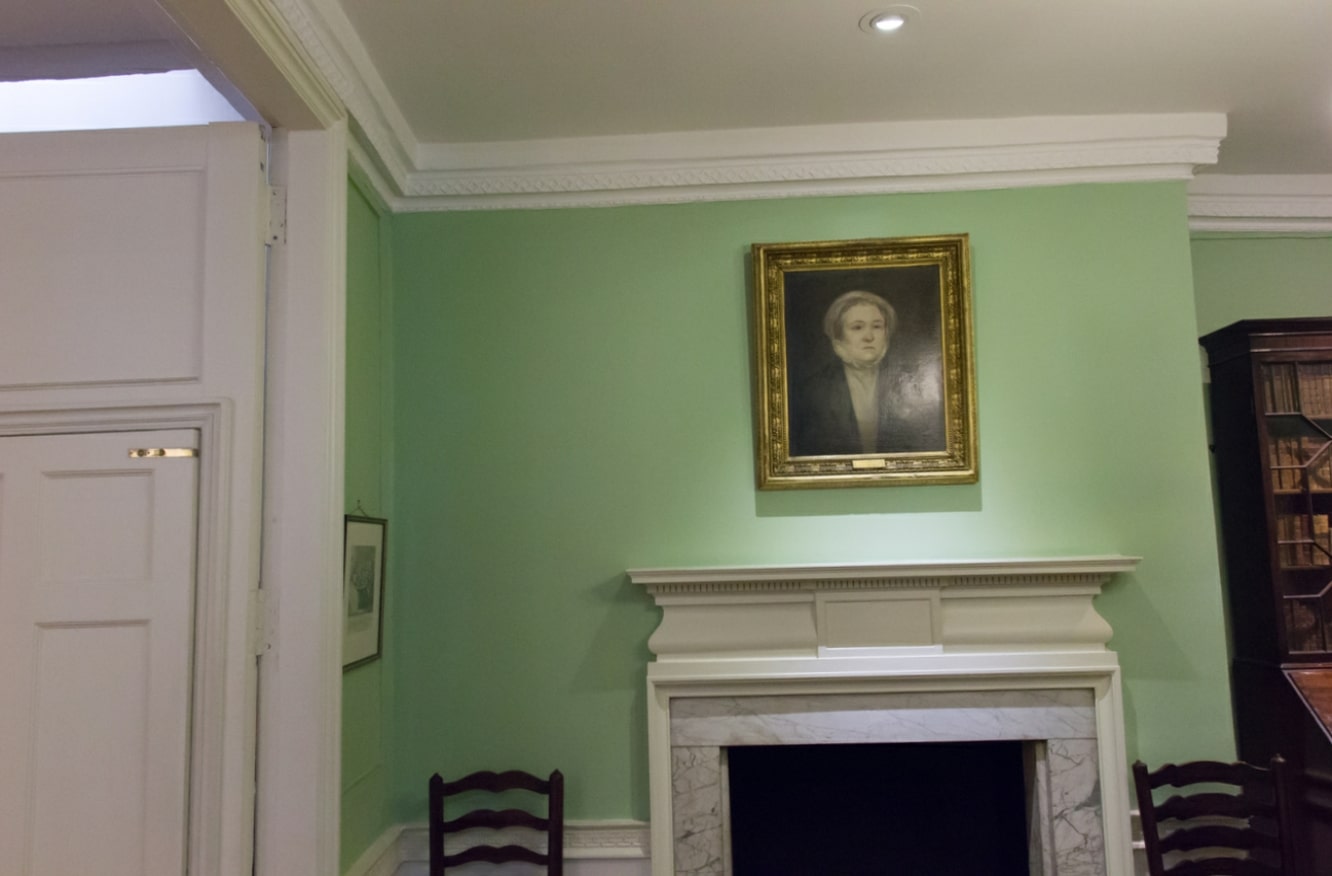}} &
        \adjustbox{valign=m}{\includegraphics[width=0.24\textwidth]{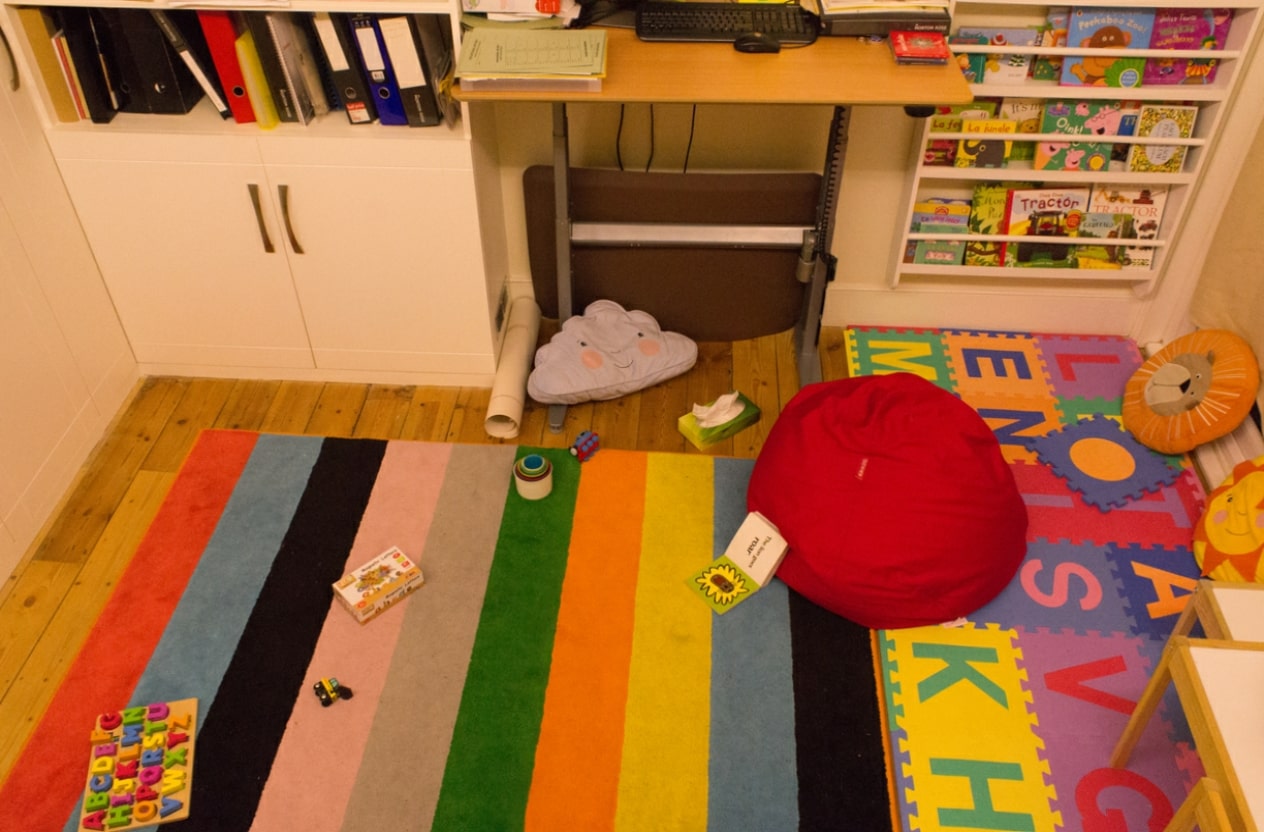}} \\
    \end{tabular}
    }
    \caption{From top to bottom are the rendering results of the following settings: 100 views selected by our method with a coarse training, all views in the dataset with a coarse training, 100 views selected by our method with a fine training, all views in the dataset with a fine training, ground truth.}
    \label{fig:rendering}
\end{figure*}

The renderings of our results are shown in Fig~\ref{fig:rendering}. The metrics evaluation of the rendering are show in Table~\ref{metrics}. Our view selection achieved similar rendering results compared to using all views from the dataset to train the 3D Gaussian models in Train, Truck and Playroom scenes. In the Dr Johnson scene, there is a gap between our results and the baseline. This is because the scene in Dr Johnson has a complex global map compared to the other scenes, so the mapping accuracy is reduced with limited image input. The reconstruction in this scene is more like a SLAM problem instead of a 3D reconstruction problem. A global planner needs to be added to improve the performance of our algorithm in this type of scenes.

\begin{table}[!htbp]
\caption{Metrics comparison of the rendering results: the best results for each scene are in bold. The second best results for each scene are underlined.}
\label{metrics}
\vspace{-0.5em}
\begin{center}
\begin{tabular}{|c|c|c|c|c|c|}
\hline
     Scene & Method & PSNR $\uparrow$ & SSIM $\uparrow$ & LPIPS $\downarrow$ \\
\hline
\multirow{4}{*}{Train} & Ours (coarse) & 18.351 & 0.7131 & 0.3106\\
                        & Dataset (coarse) & 19.546 & 0.7285 & 0.3112\\
                        & Ours (fine) & \underline{19.452} & \underline{0.7490} & \underline{0.2500}\\
                        & Dataset (fine) & \textbf{22.129} & \textbf{0.8021} & \textbf{0.2130}\\
\hline
\multirow{4}{*}{Truck} & Ours (coarse) & 17.166 & 0.6524 & 0.3664\\
                        & Dataset (coarse) & 17.060 & 0.6541 & 0.3345\\
                        & Ours (fine) & \underline{23.733} & \underline{0.8472} & \underline{0.1999}\\
                        & Dataset (fine) & \textbf{25.504} & \textbf{0.8791} & \textbf{0.1516}\\
\hline
\multirow{4}{*}{Dr Johnson} & Ours (coarse) & 21.283 & 0.7574 & 0.4418\\
                        & Dataset (coarse) & \underline{27.625} & \underline{0.8758} & \underline{0.3022}\\
                        & Ours (fine) & 22.308 & 0.7730 & 0.3756 \\
                        & Dataset (fine) & \textbf{29.232} & \textbf{0.8996} & \textbf{0.2446}\\
\hline
\multirow{4}{*}{Playroom} & Ours (coarse) & 28.475 & 0.8940 & 0.2728\\
                        & Dataset (coarse) & 28.714 & 0.8976 & 0.2688\\
                        & Ours (fine) & \underline{29.586} & \underline{0.8998} & \underline{0.2524}\\
                        & Dataset (fine) & \textbf{30.041} & \textbf{0.9023} & 0\textbf{.2444}\\
\hline
\end{tabular}
\end{center}
\end{table}

\section{CONCLUSIONS AND FUTURE WORK}

In this work, we presented a frequency-based active view selection pipeline in image acquisition process customised to 3D reconstruction in Gaussian Splatting. The pipeline significantly reduced the number of views to visit while maintaining a satisfactory quality of 3D reconstruction.

We noticed that the accuracy of camera poses and the sparse point cloud's quality significantly influence the performance of our method, as the 3D-GS's initialisation rely on both. If the camera poses of the images are not correct, the training of 3D-GS will suffer from a garbage-in-garbage-out scenario. However, based on the current SfM methods, the camera poses and sparse point cloud are difficult to estimate with only a few input images, especially in environments with complicated maps. Therefore, our method may fail in such cases, as shown in the Dr Johnson scene. We will continue to monitor the development of SfM algorithms to find a solution.

Because the training time of 3D-GS is incremented in our algorithm when the input images are added, this algorithm is not real-time in the active view selection. An option worth trying is  to select the images only close to the current camera pose for 3D-GS update, while keeping the Gaussians far from the current camera position fixed during training and rendering. Another possible solution is to utilise the parameters of the 2D Gaussians in the camera plane instead of the rendering image. As the 3D Gaussians were splatted to 2D Gaussians before the image rendering, it would more efficient to select the next view than converting them to rendered images. To that end, we would like to customize the CUDA modules to directly provide information for active view selection. Both the two possible solutions rely heavily on CUDA engineering, there are heavy workload for researchers to practice in this field.

As the current setting of the experiment is tested on datasets, the algorithm cannot select views not visited in the datasets, which limited its selection range. In our future work, we would like to deploy the algorithm in a 3D simulator and on a mobile robot to better evaluate its performance.

\addtolength{\textheight}{-12cm}   




\addtolength{\textheight}{12cm}



\section*{ACKNOWLEDGMENT}

This research was funded by by the Natural Sciences and Engineering Research Council of Canada (NSERC) via Vanier Canada Graduate Scholarships (Funding Reference Number: CGV-192714). The authors thank Dr. Dong Wang for providing the workstation with the NVIDIA GeForce RTX 4070 for the model training.




\bibliographystyle{IEEEtran}
\bibliography{main}

\end{document}